\documentclass[accepted]{uai2025} % after acceptance, for a revised version; 
\usepackage[american]{babel}
\usepackage{natbib} % has a nice set of citation styles and commands
\usepackage{mathtools} % amsmath with fixes and additions
\usepackage{booktabs} % commands to create good-looking tables
\usepackage{tikz} % nice language for creating drawings and diagrams

\usepackage{hyperref}
\usepackage{amsmath}
\usepackage{amssymb}
\usepackage{mathtools}
\usepackage{amsthm}
\usepackage[capitalize,noabbrev]{cleveref}
\usepackage{algorithm}
\usepackage{algorithmic}
\usepackage{amsfonts}
\usepackage{array}
\usepackage{bm}
\usepackage{caption}
\usepackage{color}
\usepackage{csquotes}
\usepackage{enumitem}
\usepackage{graphicx}
\usepackage{latexsym}
\usepackage{makecell}
\usepackage{microtype}
\usepackage{multibib}
\usepackage{multirow}
\usepackage{tabularx}
\usepackage{tikz}
\usepackage{times}
\usepackage{pgfplots}
\usepackage{subcaption}
\usepgfplotslibrary{groupplots}
\pgfplotsset{compat=1.17}
\usepackage{xcolor}
\usepackage{colortbl}
\definecolor{myblue}{RGB}{109,165,192}
\definecolor{myblue1}{RGB}{170,220,224}
\definecolor{myblue2}{RGB}{82,143,173}
\definecolor{blue1}{RGB}{137,201,100}
\definecolor{myred}{RGB}{231,98,84}
\definecolor{red}{RGB}{200,36,35}
\definecolor{red1}{RGB}{235,117,116}
\definecolor{red2}{RGB}{239,132,118}
\definecolor{pink}{RGB}{249,190,187}
\definecolor{myorange}{RGB}{239,138,071}
\definecolor{myyellow}{RGB}{255,208,111}
\definecolor{mygreen}{rgb}{0,0.5,0}

\title{Collapsing Sequence-Level Data-Policy Coverage \\
via Poisoning Attack in Offline Reinforcement Learning}

\author[1]{Xue Zhou}{}
\author[1]{Dapeng Man}
\author[*1]{\href{chen.xu@hrbeu.edu.cn}{Chen Xu}}
\author[1]{Fanyi Zeng}
\author[1]{Tao Liu}
\author[1]{Huan Wang}
\author[1]{Shucheng He}
\author[1]{Chaoyang Gao}
\author[1]{Wu Yang}
\affil[1]{%
    College of Computer Science and Technology, Harbin Engineering University, China
}

\begin{document}
\maketitle

\begin{abstract}
Offline reinforcement learning (RL) heavily relies on the coverage of pre-collected data over the target policy's distribution. 
Existing studies aim to improve data-policy coverage to mitigate distributional shifts, but overlook security risks from insufficient coverage, and the single-step analysis is not consistent with the multi-step decision-making nature of offline RL.
To address this, we introduce the sequence-level concentrability coefficient to quantify coverage, and reveal its exponential amplification on the upper bound of estimation errors through theoretical analysis.
Building on this, we propose the Collapsing Sequence-Level Data-Policy Coverage (CSDPC) poisoning attack. 
Considering the continuous nature of offline RL data, we convert state-action pairs into decision units, and extract representative decision patterns that capture multi-step behavior.
We identify rare patterns likely to cause insufficient coverage, and poison them to reduce coverage and exacerbate distributional shifts.
Experiments show that poisoning just 1\% of the dataset can degrade agent performance by 90\%.
This finding provides new perspectives for analyzing and safeguarding the security of offline RL.
\end{abstract}

\section{Introduction}

Offline reinforcement learning (RL) leverages pre-collected static datasets for policy learning, avoiding the high costs and risks of online exploration \citep{Levine2020OfflineRL}. 
This approach has demonstrated significant potential in domains such as robotic control \citep{Chebotar2021ActionableMU} and autonomous driving \citep{Diehl2023UncertaintyAwareMO}.

Despite offline RL's potential, its performance heavily depends on data-policy coverage \citep{Munos2003ErrorBF}, defined as how well the pre-collected data's empirical distribution matches the target policy's occupancy distribution \citep{Agarwal2019AnOP}.
Insufficient coverage exacerbates distributional shift \citep{Kumar2019StabilizingOQ}, degrading policy performance by increasing value estimation errors \citep{WangHZ23}.

Current research focuses on single-step coverage to improve algorithms, ensuring the target policy more effectively relies on available data distributions to mitigate distribution shift \citep{Fujimoto2018BCQ, kidambi2020morel}.
However, they overlook the multi-step decision-making nature of RL, where policies depend on sequences of actions rather than individual steps \citep{Sutton2018ReinforcementL}. 
Recent studies propose sequence-level methods \citep{Bar-DavidZNW23, SaanumEDBS23}, but the impact of insufficient sequence-level coverage on policy learning remains underexplored.

Building on existing research that uses the concentrability coefficient to quantify single-step coverage, we extend this concept to the sequence-level.
Through theoretical analysis, we reveal that insufficient sequence-level coverage exponentially increases the upper bound of estimation errors. 
This issue also exposes potential vulnerabilities in offline RL to adversarial attacks, attackers could exploit data poisoning to manipulate coverage and degrade agent performance.

Following this idea, we design a Collapsing Sequence-Level Data-Policy Coverage (CSDPC) attack in offline RL, employing data poisoning to reduce sequence-level coverage.
Since offline RL data is predominantly continuous, making direct coverage calculation infeasible, we first cluster the single-step state-action pairs, converting similar behaviors into unified decision units. 
Next, we extract representative decision patterns, defined as consecutive decision units with repetitions removed, to capture essential behavioral logic. 
We identify rare patterns that occur infrequently and are most likely to cause coverage insufficiency.
Finally, we minimally perturb the data to eliminate these rare patterns, significantly reducing learnable patterns and amplifying distribution shifts.
Experimental results validate the effectiveness of our approach. 
Our contributions are as follows:

\begin{itemize}
\item We analyze distributional shifts in offline RL from the perspective of data-policy coverage. Considering RL's multi-step decision-making nature, we use theory and targeted poisoning to demonstrate the critical impact of sequence-level coverage insufficiency.
\item We introduce the sequence-level concentrability coefficient to quantify coverage, and show that insufficient coverage exponentially amplifies learning errors.
\item We develop a poisoning attack to collapse data-policy coverage by precisely eliminating rare decision patterns, demonstrating that minimal perturbations can amplify distributional shifts and disrupt policy learning.
\item Experiments on multiple offline RL environments show that just 5\% perturbation magnitude on 1\% data can reduce agent performance by 90\%, and can evade existing detection methods. 
Moreover, even with only 1\% data access, the attack achieves 86\% of the effectiveness compared to full data access.
\end{itemize}

Our study reveals potential data quality issues and security vulnerabilities in offline RL under sequence-level distributional shifts, offering new directions for future research and protective measures.

\section{Related Work}
\label{Related Work}

\paragraph{Offline RL}
Due to the limitations of online RL in resources and safety \citep{levine2016end, silver2017mastering}, an increasing body of research is shifting towards offline RL, which learns policies from pre-collected datasets \citep{Levine2020OfflineRL, KumarZTL20CQLOfflineRL, FujimotoG21AMinimalist}. 

The offline RL is formalized within a Markov decision process, defined by a tuple $(\mathcal{S, A, T, R, \gamma})$, where $\mathcal{S}$ denotes the state space, $\mathcal{A}$ denotes the action space, $T (s'|s, a)$ denotes the transition distribution, $R(s, a)$ denotes the reward function, and $\gamma \in (0, 1)$ denotes the discount factor. The goal of offline RL is to find a policy $\pi(a|s)$ that maximizes the expected cumulative discounted rewards from a pre-collected dataset $D$, formalized as $\pi=arg\max\limits_{\pi}\mathbb{E}_{\tau \sim D }[\sum_{t=0}^L\gamma^tR(s, a)]$, $\tau_t=(s_t,a_t,r_t,s_{t+1})$ is a trajectory at time $t$, $L$ is the trajectory's maximum length, and $\mathbb{E}$ denotes expectation \citep{Sutton2018ReinforcementL}.

\paragraph{Data-Policy Coverage} A pivotal challenge in offline RL lies in learning when the behavioral policy's empirical distribution fails to adequately cover the target policy's occupancy distribution \citep{PrudencioASurvey}. 
Existing research leverages coverage metrics to address distributional shifts: certain studies incorporate behavioral regularization to reduce out-of-distribution actions \citep{Wu2019BehaviorRO}, while others use pessimistic penalties to constrain the policy to well-covered regions\citep{uTYEZLFM20MOPO, KidambiRNJ20MOReL}. 
However, the analysis of sequence-level coverage deficiencies in datasets and their impact on learning outcomes is still lacking. 
To better understand its effect on offline RL performance, this paper conducts a systematic study from both theoretical and empirical perspectives. 

\paragraph{Poisoning Attack}
Given that malicious attackers can manipulate decisions through erroneous data, exposing offline RL's sensitivity to such risks and exploring mitigation measures are crucial for developing robust algorithms and gaining user trust \citep{rangi2022understanding, gong2022mind}. 
Poisoning attack serves as a potent tool for evaluating algorithm vulnerabilities, and numerous studies in the online RL field have leveraged them to assist developers in gaining insights into system vulnerabilities \citep{Zhang2020RobustDR, zhang2020adaptive, Zhang2021RobustRL, SunHH21, QuSOGW21, SunZLH22, standen2023sok}. 
Moreover, some studies have employed various methods ranging from heuristic to model-driven approaches to help researchers identify weak points in RL systems \citep{lin2017tactics, kos2017delving, sun2020stealthy, yu2023learning}. 
However, these methods typically require real-time access to online training parameters or other observations, which are inapplicable in offline environments lacking interaction and synchronous updates.

As offline RL has garnered increased attention, its security challenges have come into focus. 
Early studies attacked batch RL, but the high cost of attacks at every step posed challenges \citep{ma2019policy}. 
Subsequently, researchers attempted to formulate attacks as optimization problems, devising strategies based on different costs \citep{rakhsha2020policy}. 
There are theoretical analyses of the security of offline RL \citep{rangi2022understanding}. 
Additionally, some attacks introduce triggers in datasets, conditional on the attacker's control over both training and testing stages, thus imposing stringent preconditions \citep{gong2022mind}. 

Previous studies assume attackers have elevated privileges, and neglect the impact of sequential time steps.
We propose an attack strategy to investigate the effects of sequence-level coverage insufficiency.

\section{Problem Formulation}

\subsection{Threat Model}
Following the setup in existing work \citep{ma2019policy, rakhsha2020policy,gong2022mind}, the application idea of offline RL is that anyone can serve as a data provider to openly share their experience data, and developers can use open-source data to train RL agents to reduce consumption. 

\paragraph{Attacker's Privileges}
Set the attacker as a malicious data provider or processor, capable of poisoning parts of the offline dataset. 
To ensure stealth, the poisoning rate and perturbation magnitude must be minimized to avoid detection.
Considering a more realistic scenario, we discuss the threats posed by attackers with limited data access.

\paragraph{Attacker's Goal}
The attacker's objective is to manipulate the agent by poisoning the dataset, inducing it to learn a policy that minimizes cumulative rewards or even incurs penalties, ultimately leading to erroneous actions based on the learned poisoned policy.

\paragraph{Attacker's Knowledge}
Attackers require no domain knowledge (e.g., dynamics and perception in robotics, computer vision, or vehicle dynamics in autonomous driving).
Before the attack occurs, we assume the attacker can train the agent and make an approximate estimate, a method widely used in previous work \citep{zhang2020adaptive}.

\subsection{Impact of Sequence-Level Coverage}
\label{Problem Formulation}
The performance of offline RL policy learning heavily relies on the empirical data distribution's coverage of the regions accessed by the target policy \citep{Levine2020OfflineRL}. 
Insufficient coverage can lead to significant errors in the target policy due to distributional shifts, thereby degrading policy performance.
Researchers quantified this by introducing the concentratability coefficient $C$ \citep{Munos2003ErrorBF}:
\begin{equation}
\begin{aligned}
\label{c_single}
C &=\sup_{(s,a)\in \mathcal{I}_\mu}\frac{d^\pi(s,a)}{\mu(s,a)}, \\
d^\pi(s,a)=(1-&\gamma)\,\sum_{t=0}^{\infty} \gamma^t\Pr(s_t=s,a_t=a \mid \pi),
\end{aligned}
\end{equation}
where $d^\pi(s,a)$ is the occupancy distribution under the target policy $\pi$, once $\pi$ and the environment's transition probabilities $P$ are fixed, $d^\pi(s,a)$ becomes a constant measure describing how frequently $(s,a)$ is visited by $\pi$. 
The probability $\Pr$ depends solely on $P$ and $\pi$. 
$\mu(s,a)$ denotes the empirical distribution of state-action pairs in the data collected by the behavior policy $\mu$. 
$\mathcal{I}_\mu = \{(s,a) \mid \mu(s,a) > 0\}$ ensures that $\mu(s,a)>0$ to prevent $C$ from diverging.
This ratio quantifies the disparity between the $\pi$ visitation frequency at $(s,a)$ and its coverage in the dataset.
$\sup$ represents taking the supremum over all possible $(s,a)$ pairs, it measures the worst-case distributional shift between the target policy and the empirical distribution.
Due to the limited coverage of offline datasets, if there exist any $(s,a)$ pairs such that $d^\pi(s,a) \geq \mu(s,a)$, this results in $C \geq 1$ \citep{Chen2019Concentratability}.
Moreover, when the dataset contains very sparse $(s,a)$ pairs ($\mu(s,a)$ is small) but $d^\pi(s,a)$ remains significant, $C$ can significantly increase.
This reflects insufficient coverage, which exacerbates distributional shifts, leading to inadequate policy learning and a decline in model performance \citep{Lee2020BatchRL}.

Researchers have evaluated the impact of an increased $C$ on the offline RL learning process using action-value function as an example, based on the analysis of the concentrability coefficient and error propagation from references \citep{ Munos2003ErrorBF, Chen2019Concentratability}, the upper bound of the error between the action-value function of the target policy $Q^\pi$ and the optimal $Q^{\pi^*}$ can be derived as:
\begin{equation}
\label{Q_error}
\mathbb{E}_{(s,a)\sim d^\pi}\bigl[Q^{\pi^*}(s,a)-Q^\pi (s,a)\bigr]
\le\frac{2R_{\max}C}{1-\gamma}\epsilon,
\end{equation}
where $R_{\max}$ is the maximum possible value of the reward function, and $\epsilon$ denotes the internal approximation (or regression) error under the data distribution $\mu$ (e.g., due to limited samples or function approximation), and does not directly depend on $C$. 
The equation indicates that when the data coverage is insufficient, the concentrability coefficient $C$ then serves as a shift-amplifier, mapping this internal error from $\mu$ to the target policy distribution $d^\pi$, ultimately enlarging the Q function estimation bias.

Although the concentrability coefficient is instrumental in enhancing policy learning \citep{WangHZ23}, existing studies remain constrained to measuring deviations at individual state-action pairs, overlooking the multi-time-step decision-making nature of reinforcement learning.
Recent studies have increasingly focused on sequence-level analysis \citep{Bar-DavidZNW23, SaanumEDBS23}. 
To better characterize the distributional properties of offline RL data, we extend the concentrability coefficient to the sequence level.
Sequence-level concentrability coefficient $C_\tau$ is defined as:
\begin{equation}
C_\tau = \sup_{\tau \in \mathbb{T}_\mu} \frac{d^\pi(\tau)}{\mu(\tau)},
\end{equation}
where $\tau=(s_0,a_0),(s_1,a_1),\dots,(s_{l-1}, a_{l-1})$ is a sequence of $l$ time steps, $\mu(\tau)$ denotes the probability distribution of this sequence in the offline data, and $\mathbb{T}_\mu = \{\tau \mid \mu(\tau) > 0\}$ ensures $\mu(\tau)>0$ to prevent the ratio remains finite, and $d^\pi(\tau)$ represents the probability distribution of generating this sequence under the target policy. 
By decomposing the trajectory probabilities, we obtain:
\begin{equation}
\begin{aligned}
\frac{d^\pi(\tau)}{\mu(\tau)}=& 
\frac{\prod_{t=0}^{l-1} \big[\pi(a_t \mid s_t) P(s_{t+1} \mid s_t, a_t)\big]}
{\prod_{t=0}^{l-1} \big[\mu(a_t \mid s_t) P(s_{t+1} \mid s_t, a_t)\big]}\\
=&\prod_{t=0}^{l-1}\frac{\pi(a_t\mid s_t)}{\mu(a_t\mid s_t)}.
\end{aligned}
\end{equation}

\begin{figure*}[t!]
\begin{center}   
\centerline{\includegraphics[width=1\linewidth]{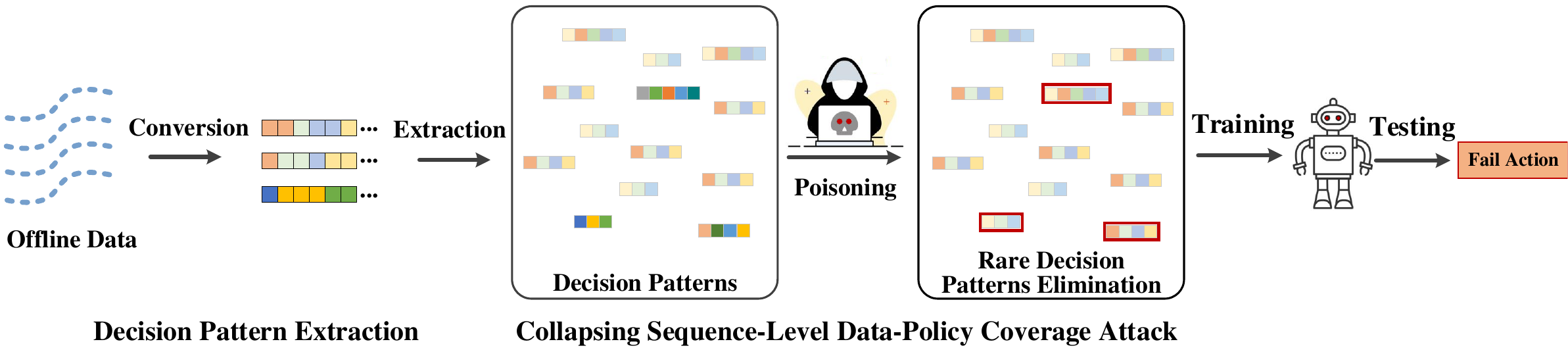}}
\caption{The workflow of the CSDPC. We first convert continuous data into decision units and merge consecutively repeated units to extract representative decision patterns.
Next, we poison the dataset to transform rare patterns into common ones, thereby reducing coverage and exacerbating distributional shifts, leading to a decline in the performance of poisoned agents.}
\label{fig: CSDPC Attack}
\end{center}
\end{figure*}

According to Equation \ref{c_single}, $C$ is the supremum, implying that for each time step $\tfrac{\pi(a_t|s_t)}{\mu(a_t|s_t)} \le C$, and assuming the transition probabilities $P$ are consistent under both the behavior and target policies \citep{Sutton2018ReinforcementL}, then: 
\begin{equation}
C_\tau=\sup_{\tau \in \mathcal{T}_\mu}\frac{d^\pi(\tau)}{\mu(\tau)}=\sup_{\tau \in \mathcal{T}_\mu}\prod_{t=0}^{l-1}\frac{\pi(a_t\mid s_t)}{\mu(a_t\mid s_t)}\le \prod_{t=0}^{l-1}C=C^l.
\end{equation}
This suggests that with insufficient coverage of some sequence, the sequence-level concentrability coefficient $C_\tau$ can reach up to the $l$ power of the single-step coefficient $C$ ($C \geq 1$). 
We further quantify Q-value estimation errors at the sequence level.
Given that value estimation in RL incorporates discount weights over time steps to reflect the decay of future rewards or errors \citep{Sutton2018ReinforcementL}, we adopt a discounted weighted sum when measuring sequence-level Q-value errors. 
Based on Equation \ref{Q_error}, the weighted Q-value estimation error at the sequence level is:
\begin{equation}
\begin{aligned}
\mathbb{E}_{\tau \sim d^\pi}&
\Bigl[\sum_{t=0}^{l-1} \gamma^t \bigl(Q^{\pi^*}(s_t,a_t) - Q^{\pi}(s_t,a_t)\bigr)\Bigr]\\
\le& \frac{2R_{\max}C_\tau}{1-\gamma}\epsilon \le \frac{2R_{\max}C^l}{1-\gamma}\epsilon.
\end{aligned}
\end{equation}
This shows that when certain sequences occur infrequently ($\mu(\tau)$ is small),  the increase in $C_\tau$ indicates insufficient coverage, leading to exponential growth in value estimation error bounds and significantly impacting the performance of the target policy. 
Altering the distribution of these rare sequences can exacerbate distributional shifts, further amplifying estimation errors and degrading policy performance, which offers a new perspective on data poisoning.

\section{Method}

The above analysis indicates that insufficient sequence-level coverage exponentially amplifies the upper bound on the estimation error of the $Q$ function.
To further verify the impacts, we design the Collapsing Sequence-Level Data-Policy Coverage (CSDPC) poisoning attack to reduce sequence-level coverage.

In realistic offline RL scenarios, such as robotics and autonomous driving \citep{Fu2020D4RLDF}, high-dimensional continuous data make it infeasible to compute the sequence-level coverage directly, so we first converse state-action pairs into decision units to enable coverage analysis (Section \ref{Partitioning Decision Units}).
Next, we extract decision patterns from sequences by removing repetitions, compactly representing multi-step behavioral logic (Section \ref{Decision Pattern Extraction}). 
We then identify rare decision patterns that correspond to low-coverage regions and are most likely to amplify distributional shifts and estimation errors (Section \ref{Rare Decision Pattern Identification}).
Finally, we inject minimal perturbations to eliminate these patterns, reducing sequence-level coverage (Section \ref{Perturbing Decision Sequences}).
Figure \ref{fig: CSDPC Attack} is the CSDPC attack workflow, and detailed steps in Appendix \ref{CSDPC Algorithm}.

\subsection{Decision Unit Conversion}
\label{Partitioning Decision Units} 

To address the challenge of handling continuous data in offline RL, a natural idea is to discretize continuous state-action pairs into countable units. 
The goal is to map each $(s, a) \in \mathcal{S} \times \mathcal{A}$ to a discrete unit $u_j \in U$, where $U={u_1,u_2,\dots,u_k}$ is a finite set of decision units. 
This mapping can be formalized as:
\begin{equation} 
\label{discretization} 
\phi(s_t, a_t) = \text{Discretize}(s_t, a_t),
\end{equation}
where $\phi:\mathcal{S} \times \mathcal{A} \rightarrow U$ is the discretization function. 
We employ the classical k-means clustering algorithm to achieve this discretization, and the optimal number of clusters $k$ is determined using the elbow method \citep{Jain2008DataC5}, each cluster is represented by a unit $u_j$.

However, previous studies have emphasized that high-dimensional raw inputs usually contain redundant or noisy information  \citep{Mnih2015HumanlevelCT}. 
Therefore, direct clustering of raw $(s, a)$ may not be effective in capturing potential decision-making behaviors.
Some research suggests that higher-level feature representations derived from raw data better capture the essential decision-making logic of RL agents \citep{Fujimoto2018BCQ, Kumar2020CQL}. 

Inspired by this, we train an RL agent on a clean dataset $D$, and use its trained encoder network to extract meaningful features from the raw data. 
Specifically, for each time step, we encode the state-action pair $(s, a)$ into a feature vector $f_t$.
Formally, Equation \ref{discretization} can be formalized as:
\begin{equation}
\label{clustering_features}
\phi(f_t) = \text{K-means}[\text{Encoder}(s, a)].
\end{equation}
The encoder maps the raw state-action pair to a higher-level, decision-relevant feature representation.
Compared to clustering raw data, this method provides a more precise and agent-aware approach to decision unit conversion.

\subsection{Decision Pattern Extraction}
\label{Decision Pattern Extraction}

After converting state-action pairs into decision units, the next step is to extract sequence-level decisions from sequences to compute coverage.

Given a trajectory of length $L$, from which we extract a sequence $\tau_t = (s_t, a_t), (s_{t+1}, a_{t+1}), \ldots, (s_{t+l}, a_{t+l})$ with a sequence length $l$ satisfying $2 \leq l \leq L$, where $t$ denotes the starting index. 
After discretizing the state-action pairs into decision units, each $(s, a)$ is mapped to a clustering label $u \in \{1, 2, \ldots, k\}$, resulting a sequence of behaviors represented by the cluster labels:
\begin{equation}
\tau_t = u_t, u_{t+1}, \ldots, u_{t+l}.
\end{equation}

The decision space for sequences of length $l$ is too large by $k^l$, thus complicating the analysis.
Meanwhile, consecutive decision units often represent similar behaviors. 
We merge consecutively repeated units to obtain the de-duplicated consecutive decision units, defined as the decision pattern $p$. 

This step ensures the representation highlights meaningful behavioral changes while reducing redundant repetition.
This deduplication strategy is particularly suitable for continuous or near-continuous control tasks (e.g., robotics, autonomous driving), where repetitive actions dominate most time steps. 
For instance, straight-line driving occurs frequently, potentially overshadowing sparse yet crucial behaviors such as turning. 
Deduplication helps compress redundant behaviors and better reflect the structure of meaningful decision patterns.

\subsection{Rare Decision Pattern Identification}
\label{Rare Decision Pattern Identification}

Theoretical analysis indicates that a higher sequence-level concentrability coefficient $C_\tau$ corresponds to lower data-policy coverage, thereby exacerbating distributional shifts.
To exploit this vulnerability, our attack aims to increase $C_\tau$, thereby effectively reducing coverage and amplifying the impact of distributional shifts.
The most effective approach is to decrease the frequency of rare sequences, as these sequences are most likely the contributors to low-coverage regions.
Thus, it is necessary to identify rare sequences.
We previously extracted representative decision
patterns from sequences. 
Next, we identify rare decision patterns.

To identify rare decision patterns, we first extract decision patterns $p$ from all sequences in the dataset, and compute the occurrence frequency of each pattern denoted as $O(p)$. 
Next, we rank all patterns by their occurrence frequency $O(p)$.
Based on a limited poisoning rate $\rho$ ($0 < \rho < 1$), selecting the patterns with the lowest frequencies to form the rare decision pattern set $\mathcal{P}$. 

These rare patterns correspond to the sequence-level sparse regions identified in the theoretical analysis and most likely lead to an increase in $C_\tau$. 
The rare decision pattern set $\mathcal{P}$ serves as the target for subsequent poisoning operations.

\subsection{Coverage Reduction via Poisoning}
\label{Perturbing Decision Sequences}

To achieve the object of reducing sequence-level data-policy coverage, the attack must ensure both effectiveness and stealthiness, emphasizing imperceptibility to simulate real-world adversarial scenarios.

Specifically, we devise a perturbation strategy based on two crucial components: First, introducing a stealthiness constraint for the perturbation magnitude to ensure the attack proceeds without causing conspicuous anomalies to evade detection;
Second, we reduce coverage by erasing rare decision patterns from the dataset through poisoned perturbations, thereby increasing $C_\tau$.

\paragraph{Stealth Constraint Imposition}

Our objective is to explore a more stealth setup that minimizes the attack's magnitude and adapts it to the size of the original data. 
We introduce a perturbation $\zeta_t$, generating the poisoned state-action pair $(s_t+\zeta_t^s, a_t+\zeta_t^a)$. The perturbation $\zeta_t$ is subject to the following constraints: 
\begin{equation}
    \parallel \zeta_t^s \parallel_{\infty} < \eta \cdot \parallel s_t \parallel_{\infty}\;\;, \;\;\parallel \zeta_t^a \parallel_{\infty} < \eta \cdot \parallel a_t \parallel_{\infty},
\end{equation}
where $\eta$ represents the perturbation ratio, almost set to 0.05 in our paper. 
This ensures that the perturbations remain a small portion of the original values, thus maintaining the stealth of the attack.

\begin{table*}[t!]
  \centering
\caption{The ACR obtained by the agent in various environments, the values in parentheses are the AER.}
  \label{4environments_table}
  \renewcommand{\arraystretch}{1.3}
  \resizebox{\textwidth}{!}{
  \begin{tabular}{c|rrr|rrr|rrr|rrr}
    \toprule
     \multirow{2}{*}{Algorithms} &\multicolumn{3}{c|}{Walker2D} &\multicolumn{3}{c|}{Hopper}&\multicolumn{3}{c|}{Half}&\multicolumn{3}{c}{Carla}\\ 
    \cline{2-13} 
    &\multicolumn{1}{c}{Clean}&\multicolumn{1}{c}{Raw}&\multicolumn{1}{c|}{Feature}&\multicolumn{1}{c}{Clean}&\multicolumn{1}{c}{Raw}&\multicolumn{1}{c|}{Feature}&\multicolumn{1}{c}{Clean}&\multicolumn{1}{c}{Raw}&\multicolumn{1}{c|}{Feature}&\multicolumn{1}{c}{Clean}&\multicolumn{1}{c}{Raw}&\multicolumn{1}{c}{Feature} \\ 
    \hline
    CQL & 3132 &438(86\%) & 263(92\%) & 3158 & 380(88\%) & 234(93\%)& 4822& 626(87\%) & 513(89\%) & 191 & 61(68\%) & 30(84\%)\\
    BEAR& 2593 & 221(91\%) & 172(93\%) & 2119 & 215(90\%) &131(94\%)&  4290& 516(88\%) & 421(90\%) & 89 & 26(71\%) & 11(87\%)\\
    BCQ & 2341 & 365(84\%) & 223(90\%) & 2823 & 280(89\%) &203(93\%)& 4694& 904(81\%) & 772(84\%) & 466 & 153(67\%) & 72(85\%)\\ 
    BC & 744 & 107(86\%)& 62(92\%) & 3450 & 384(89\%) & 226(93\%) & 4017& 516(87\%) & 400(90\%) & 384 & 128(67\%) & 70(82\%)\\
    \cline{1-13} 
    \multirow{2}{*}{Average} & 2203 & 285(87\%) & 161(92\%) & 2613 & 315(89\%) & 199(93\%) & 4456 & 641(86\%) & 527(88\%) & 283 & 92(68\%) & 46(85\%)\\
    \cline{2-13} 
     & \multicolumn{6}{c}{Raw Data AER \quad 83\%} & \multicolumn{6}{c}{Feature Data AER \quad 90\%}\\
    \bottomrule    
  \end{tabular}}
\end{table*}

\paragraph{Rare Decision Pattern Elimination}

To effectively reduce sequence-level coverage, our attack focuses on transforming the rarest decision patterns into the most frequent ones through targeted poisoned perturbations. 
By increasing the occurrence of these rare patterns, thereby significantly minimize sequence-level coverage.

For each raw sequence $\tau$ corresponding to a rare decision pattern $p\in \mathcal{P}$, we generate multiple potential poisoned sequences $\tau^1,\tau^2,\dots,\tau^n$ under a stealth constraint. 
These multiple potential poisoned sequences ensures that the perturbation most capable of replacing the raw rare pattern can be identified, thereby maximizing the attack's impact while adhering to the stealth constraint.
Each perturbed sequence undergoes clustering and extraction operations to obtain corresponding decision patterns $p^1, p^2, \ldots, p^n$. We then compute the occurrence counts $O_{p^i}$ for each poisoned decision pattern in the dataset.

To maximize the attack's impact, we select the perturbed sequence $\tau^i$ that results in the highest occurrence count $O_{p^i}$, as it most effectively replaces the raw rare decision pattern. 
This selected poisoned sequence $\tau^n$ replaces the raw sequence $\tau$.
Applying this poisoning process to all data results in the poisoned dataset $D'$.

This replacement operation significantly reduces the presence of rare decision patterns in the dataset, thereby lowering sequence-level coverage. 
As a result, the target policy is forced to rely on a limited set of decision patterns, exacerbating distributional shifts.

\section{Experiments and Analysis}
\label{Experiments and Analysis}

In this section, we conduct a comprehensive ablation study on the Collapsing Decision Pattern Diversity (CSDPC) attack.
We evaluate these attacks under various offline RL settings across diverse environments and provide an in-depth analysis of the experimental results. 

\subsection{Experimental Settings}

\paragraph{Settings} Considering stealthiness, we set our poisoning proportion range to $\rho=\{1\%, 5\%, 10\%, 20\%\}$, with most experiments were conducted at $\rho=1\%$ and $\rho=5\%$. 
The perturbation magnitude ranged as $\eta=\{0.05, 0.15, 0.25\}$, where most of the experiments were performed with constraints of $\eta=0.05$.
Unless specified otherwise, the attacks mentioned in the experimental section are based on the raw data.
The detailed experimental settings is outlined in Appendix \ref{Experimental Settings}. 

\paragraph{Evaluation Metrics} The Average Cumulative Reward (ACR), defined as $R = \frac{1}{T} \sum_{t=0}^T r_t$, directly measures the agent's task performance in the test environment.
In our experiments, $T$ is set to 50, meaning the ACR reflects the agent's average performance across 50 trajectories in the test environment.

To evaluate the impact of different attack methods on agent performance, we adopt the commonly used metric: attack effectiveness rate (AER), which quantifies the performance degradation of the poisoned agent compared to the clean agent. AER is calculated as:
\begin{equation}
\text{AER} = \frac{R_{clean}-R'_{poisoned}}{R_{clean}} * 100\%,
\end{equation}
where $R'_{poisoned}$ is the average cumulative reward of the poisoned model in the trigger environment. 
A higher AER indicates greater attack effectiveness.

\subsection{Overall Result of CSDPC}

\paragraph{Effectiveness}
Table \ref{4environments_table} shows the attack outcomes in four tasks with a mere 1\% poisoning rate, where CSDPC attack can reduce the agents' average cumulative rewards (ACR) by 83\% when partitioning the decision space using raw data, as effective as the 10\% poisoning rate in BAFFLE \citep{gong2022mind}. 
Moreover, when attackers can extract advanced features from the data, the attacks can diminish the agents' average performance by 90\%. 
The $\rho=5\%$ and more comparative results are provided in Appendix \ref{More Results}. 

These results show that our attack can significantly disrupt offline RL algorithms across various tasks with minimal cost. 
The effectiveness of the attack increases with the attacker's knowledge, indicating that the attacker can capture more crucial information during the learning process, thereby hindering the agent from learning optimal policies.

\paragraph{Generalizability}
Although our analysis is based on Q-values, the results in Table \ref{4environments_table} reveal that the CSDPC attack remains effective against the BC algorithm, which does not use Q-values updates. 
Additionally, CSDPC attacks significantly degrade agent performance in complex tasks across different data types (e.g., numeric inputs for Mujoco robot tasks, and image inputs for Carla autonomous driving tasks).
We further conduct additional experiments on AWAC \cite{kumar2020awac}, a policy gradient method. 
Even $\rho=1\%$ and encoder mismatch, the AER is 85\% in Walker2D.
These results indicate that reducing the coverage of rare decision patterns has a significant impact on various offline RL algorithms for a wide range of tasks.
This is attributed to the fact that our attack method does not rely on any specific offline RL algorithm or task structure design, but rather impacts agents by targeting the essence of coverage.

\begin{table*}[t!]
  \centering
\caption{CSDPC attacks based on raw data were executed on the CQL agent in the Walker2D task under varying data privileges and perturbation magnitudes. Small-range data was randomly selected as consecutive time steps from the dataset.}
  \label{less_privileges}
  \renewcommand{\arraystretch}{1.3}
  \resizebox{\textwidth}{!}{
  \begin{tabular}{c|c|rrrrr|rrrrr}
    \toprule
     \multirow{2}{*}{Clean} & \multirow{2}{*}{$\eta$} &\multicolumn{5}{c|}{$\rho=1\%$} & \multicolumn{5}{c}{$\rho=5\%$}\\ 
    \cline{3-12} 
    &&\multicolumn{1}{c}{Whole Data}&\multicolumn{1}{c}{20\% Data}&\multicolumn{1}{c}{10\% Data}&\multicolumn{1}{c}{5\% Data}&\multicolumn{1}{c|}{1\% Data}&\multicolumn{1}{c}{Whole Data}&\multicolumn{1}{c}{20\% Data}&\multicolumn{1}{c}{10\% Data}&\multicolumn{1}{c}{5\% Data}&\multicolumn{1}{c}{1\% Data} \\ 
    \hline
    \multirow{3}{*}{3132}& 0.05 & 438(86\%) & 604(81\%) & 665(79\%) & 725(77\%) & 952(70\%)& 322(90\%) & 489(84\%) & 541(83\%) & 637(80\%) & 856(73\%)\\
    & 0.15& 361(88\%) & 530(83\%) & 564(82\%) & 680(78\%) & 905(71\%) & 213(93\%) & 438(86\%) & 478(85\%)  & 542(83\%) & 737(76\%)\\
    & 0.25 & 299(90\%) & 433(86\%) & 527(83\%) & 620(81\%) & 723(77\%) & 184(94\%) & 384(88\%) & 402(87\%) & 499(84\%) & 622(80\%)\\ 
    \bottomrule    
    \end{tabular}}
\end{table*}

\begin{table}[t!]
    \centering
    \caption{Anomaly value-based detection results, $\rho=5\%$.}
    \label{tab: AnomalyDetection}
    \renewcommand{\arraystretch}{1.2} % Adjust row height
    \footnotesize
    \setlength{\tabcolsep}{3pt} % Adjust column separation
    \begin{tabular}{c|cc|cc|cc}
        \toprule
        \multirow{2}{*}{Environments} & \multicolumn{2}{c|}{Precision} & \multicolumn{2}{c|}{Recall} & \multicolumn{2}{c}{F1-score} \\
        \cline{2-7}
        &clean&poison&clean&poison&clean&poison\\
        \hline
        Hopper &8.0\% & 7.6\% &24.3\% & 24.5\% &11.6\% & 12.8\% \\
        Half &7.9\% & 9.0\% &24.1\% & 25.1\% &13.5\% & 12.7\% \\
        Walker2D &6.9\% & 6.9\% &25.8\% & 24.8\% &10.9\% & 11.9\% \\
        \hline
        Average &7.6\% & 7.8\% &24.7\% & 24.8\% &12.0\% & 12.5\% \\
        \bottomrule
    \end{tabular}
\end{table}

\paragraph{Stealthiness}
To enhance the stealthiness of the attack, we explore the effectiveness of attacks when the attacker has only limited data upload or processing privileges. 
As demonstrated in Table \ref{less_privileges}, even with access to as little as 1\% of the data, the attacker can significantly degrade the agent's performance (e.g., reducing returns by 70\% under $\eta=0.05, \rho=1\%$). 
Moreover, the attack's effectiveness consistently increases as the attacker gains access to a larger privileges of the data. 
This trend suggests that having access to more data allows the attacker to identify and perturb a greater number of critical decision points within the agent's learning process, leading to more substantial damage. 

We also investigate the impact of varying perturbation magnitudes on the attack effectiveness. 
We observe that larger perturbations can generate poisonous samples that reside in a different cluster from the raw data, eradicating more rare decision patterns, thus significantly reducing the rare decision pattern coverage in the dataset, as reflected in the results in Table \ref{less_privileges}. 
Another clear trend is that larger perturbation sizes can offset the diminished effectiveness of attacks due to restricted data access. 
For example, when the perturbation size is increased from 0.5 to 0.25, with the same poisoning rate, 5\% data access can achieve the same attack effectiveness as with 20\% data access.

To further evaluate the stealthiness of the CSDPC attack, we employed GradCon \citep{KwonPTA20/ECCV}, a widely used anomaly detection approach, to detect clean and poisoned data points. 
We generated poisoned data by injecting poison into the Walker2D dataset using the CQL algorithm with the CSDPC attack, $\rho=5\%$. 
The results in Table \ref{tab: AnomalyDetection} indicated no significant anomalies in the detection of poisoned data compared to clean data. 
Notably, the average F1 score for poisoning was only 12.5\% using our method, in contrast to GradCon's over 87\% F1 score on other datasets. 
This indicates that the anomalies were almost undetectable, thereby demonstrating the stealthiness of our approach.

\subsection{Ablation Study}

\paragraph{Continuous vs. Discrete Poisoning}
To investigate whether continuous time steps have a greater impact on the learning process compared to discrete ones, we identify critical sequences of length 10 based on evaluation criteria. 
We perturb a random sequence of length 5 and discrete time steps spanning 5 intervals within each critical sequence using the CSDPC attack. 
The results in Table \ref{sequences_timesteps} indicate that within the same range, attack sequences lead to a greater degradation in the agent's performance, thus demonstrating a more significant impact on the learning process in continuous timesteps compared to discrete ones.

\begin{table}[t!]
  \centering
\caption{Perturbing consecutive and discrete time steps within the same sequence of CQL agents trained on the Walker2D dataset under full data privilege.}
  \label{sequences_timesteps}
  \renewcommand{\arraystretch}{1.2}
  \footnotesize
  \resizebox{0.48\textwidth}{!}{
  \begin{tabular}{c|cc|cc}
    \toprule
    \multirow{2}{*}{Data} &\multicolumn{2}{c|}{$\rho=1\%$}&\multicolumn{2}{c}{$\rho=5\%$}\\ 
     \cline{2-5} 
     &Sequence& Discrete Steps & Sequence& Discrete Steps\\
    \hline
     Raw& 478(85\%) & 539(83\%) & 372(88\%) & 443(86\%) \\
     Feature& 377(88\%) & 419(87\%) & 243(92\%) & 348(89\%) \\
    \bottomrule    
  \end{tabular}}
\end{table}

\begin{table}[t!]
  \centering
\caption{Attack results under different $n$-step CQL configurations in Walker2D.}
  \label{nstep_cql}
  \renewcommand{\arraystretch}{1.2}
  \scriptsize
  \resizebox{0.45\textwidth}{!}{
  \begin{tabular}{c|c|c|c|c}
    \toprule
Algorithm & $\rho$ & n-step & Clean & ACR(AER) \\
\hline
\multirow{2}{*}{CQL} & \multirow{2}{*}{$1\%$} & 1 & 3132  & 438(86\%) \\
& & 5 & 3296 & 297(91\%) \\
\bottomrule    
  \end{tabular}}
\end{table}

In addition, to assess the impact of multi-step corruption, we evaluate our attack on CQL with $n$-step = 5, which explicitly incorporates multi-step return estimation. 
As shown in Table~\ref{nstep_cql}, the agent experiences even greater performance degradation than the single-step version under the same poisoning budget, supporting our theoretical insight that sequence-level coverage insufficiency has a more pronounced effect in multi-step algorithms.
These findings confirm that the more an algorithm relies on sequence information, the more vulnerable it becomes to sequence-level poisoning, further validating the practical relevance of our attack.

% 第一个图
\begin{figure}[t!]
    \centering
    \begin{tikzpicture}
        \begin{axis}[
            ybar,
            ymajorgrids,
            xmajorgrids,
            grid style=dashed,
            symbolic x coords={Random, C-value, Q-value, CSDPC}, 
            xtick=data,
            enlarge x limits=0.12, % 缩小边界扩大比例
            xtick style={draw=none},
            xmin=Random,
            xmax=CSDPC,
            xticklabel style={font=\footnotesize, yshift=2pt},
            ytick style={draw=none},
            axis line style={draw=black},
            ymin=0, ymax=1000,
            ytick={0,250,500,750,1000},
            yticklabel style={font=\footnotesize},
            ylabel style={yshift=-0.2cm},
            ylabel=\footnotesize{ACR},
            bar width=0.4cm, % 根据需要调整柱状图宽度
            width=0.95\linewidth, % 确保图形宽度适应当前栏
            height=3.6cm % 根据需要调整高度
        ]
        \addplot[draw=none, fill=myblue] coordinates {
            (Random, 910.4351)        
            (C-value, 793.2804)
            (Q-value, 839.1451)            
            (CSDPC, 438.2528)
        };
        \end{axis}
    \end{tikzpicture}
    \caption{Identify critical sequences methods.}
    \label{fig:Method_For_Identify_Sequences}
\end{figure}
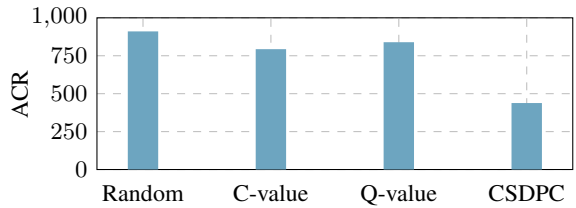

\paragraph{Identifying Sequences}
To discuss the impact of sequences identified by different evaluation criteria on the attack, we utilize random selection, C-value \citep{lin2017tactics}, Q-value \citep{kos2017delving}, and our CSDPC attack method to determine critical sequences of length 5.  
Utilizing the perturbation method under constraints, we introduce perturbations with magnitudes within 0.05 to state-action pairs of the selected critical sequences. 
Figure \ref{fig:Method_For_Identify_Sequences} demonstrates the effectiveness of attacking different sequences.
Results indicate that our method can accurately select sequences with a greater impact on the learning process.

This is may because high-value sequences often correspond to prevalent decision patterns (e.g., fast straight-line actions in autonomous driving tasks). 
As these patterns occur frequently in the dataset, the agent's learning of them is more stable, resulting in a lesser impact on the agent's performance when these patterns are perturbed.
Conversely, CSDPC attack results in the absence of rare decision patterns within the dataset, considerably impairing the agent's capability in scenarios of inadequate learning, and subsequently precipitating a significant drop in performance.

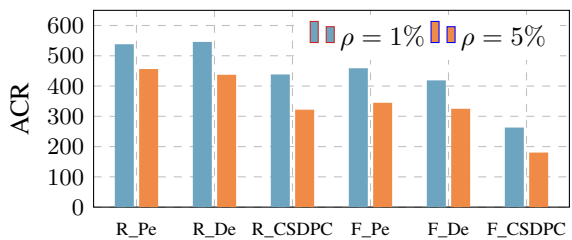
\begin{figure}[t!] 
    \centering
    \begin{tikzpicture}
    \begin{axis}[
        ybar, % 条形图选项
        ymajorgrids,
        xmajorgrids,
        grid style=dashed,
        bar width=.25cm, % 条的宽度
        width=0.95\linewidth, % 确保图形宽度适应当前栏
        height=4.2cm, % 根据需要调整高度
        enlarge x limits=0.25, % x轴的边界扩大
        legend style={at={(0.98,0.98)},
        anchor=north east,legend columns=-1,
        draw=none}, % 图例设置
        symbolic x coords={R\_Pe, R\_De, R\_CSDPC, F\_Pe,  F\_De, F\_CSDPC}, % x轴坐标
        xtick=data, % 以数据点为刻度标记
        enlarge x limits=0.11,
        xtick style={draw=none},
        ytick style={draw=none},
        xticklabel style={font=\scriptsize, yshift=2pt},
        yticklabel style={font=\footnotesize},
        ylabel style={yshift=0pt},
        ylabel={ACR}, % Y轴标签
        nodes near coords align={vertical}, % 数值对齐方式
        ymin=0,ymax=650, % Y轴范围
        ytick distance=100, % Y轴刻度距离
        cycle list name=color list,% 色彩列表
        ]
    \addplot+[ybar, fill=myblue, draw=none] coordinates {
    (R\_Pe, 538.1068) 
    (R\_De, 545.726)
    (R\_CSDPC,438.2528)
    (F\_Pe, 458.7503) 
    (F\_De, 418.6589)
    (F\_CSDPC, 262.8584)
    };
    \addplot+[ybar, fill=myorange, draw=none] coordinates {
    (R\_Pe, 456.1134) 
    (R\_De, 437.1937)
    (R\_CSDPC, 321.9049)
    (F\_Pe, 344.6785)
    (F\_De, 324.8855)
    (F\_CSDPC, 180.2488)
    };
    \legend{$\rho=1\%$,$\rho=5\%$}
    \end{axis}
    \end{tikzpicture}  
    \caption{Different methods of poisoning.}
    \label{fig:reduceDSC}
\end{figure}

\paragraph{Poisoning Methods}
To explore the effect of eliminating rare decision patterns on attacks, we compare the CSDPC attack method proposed in this paper with an approach that merely introduces perturbations under certain constraints (denoted as "Pe"), and those that directly delete sequences corresponding to rare decision patterns (denoted as "De"). 
Results depicted in Figure \ref{fig:reduceDSC} indicate that the CSDPC attack is more effective under various poisoning rates and levels of attacker knowledge (where "R" represents the use of raw data, and "F" indicates the use of advanced features). 
Compared to these two methods, the CSDPC attack not only eliminates rare decision patterns but also alters the distribution of other common decision patterns, thereby exerting a greater negative impact on the agent.

\begin{table}[t]
\centering
\caption{Impact of deduplication on attack performance, $\rho=1\%$.}
\label{tab:dedup_ablation}
\renewcommand{\arraystretch}{1.2}
\footnotesize
\begin{tabular}{r|c|c|c}
\toprule
\multicolumn{1}{c|}{Setting} & Clean & ACR & AER \\
\hline
w/o Deduplication & \multirow{2}{*}{3132} & 971 & 69\% \\
w/ Deduplication    && 438 & 86\% \\
\bottomrule
\end{tabular}
\end{table}

\paragraph{Effect of Deduplication}
To evaluate the impact of our deduplication strategy, we conducted additional experiments on the Walker2D dataset. 
Compared to the original sequences, deduplication reduced the number of distinct decision patterns by nearly 80\%, highlighting its role in removing repetition and preserving meaningful behavior changes.
We further evaluated the impact on attack effectiveness using the CQL algorithm under a 1\% poisoning rate with raw data. 
As shown in Table~\ref{tab:dedup_ablation}, deduplication significantly improves the attack's effectiveness.

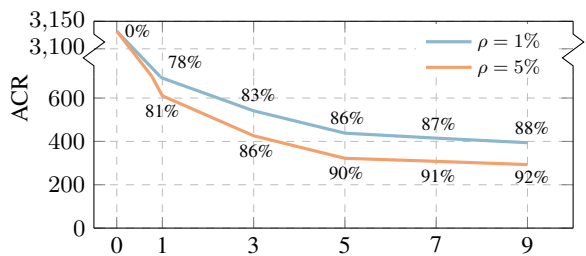
\begin{figure}[t!] % 使用figure环境来确保表格和图片都能被正确引用  
\centering  
    \begin{tikzpicture}   
    \footnotesize{
        \begin{groupplot}[
        ymajorgrids,
        xmajorgrids,
        grid style=dashed,
        legend style={
        draw=none,
        line width=1pt,
        },
        legend style={at={(0.96,0.98)}, anchor=north east, legend cell align=left,
        nodes={scale=0.8, transform shape}},
        xmode=normal,
        group style={
                group name=my fancy plots,
                group size=1 by 2,
                xticklabels at=edge bottom,
                vertical sep=0pt
            },
        xmin=-0.5,xmax=10,
        xtick distance=1,
        xtick={0,1,3,5,7,9},
        xticklabels={0,1,3,5,7,9},
        xlabel style={yshift=0.0em},
        ]        \nextgroupplot[ymin=3050,ymax=3150,
                       ytick={3100,3150},
                       axis x line*=top, 
                       ylabel style={at={(-0.12, -0.4)}}, 
                       ylabel=\footnotesize{ACR},
                       axis y discontinuity=crunch,
                       width=0.46\textwidth,
                       height=0.135\textwidth]              
    \addplot[myblue!80, line width=1.2pt] plot coordinates {
        (0,3132)
        (0.95,3050)
        };    
        \node[right,font=\scriptsize] at (axis cs:0.001,3132) {0\%};
        \node[above,font=\scriptsize] at (axis cs:1.5,3050) {78\%};

        \addlegendentry{$\rho=1\%$}
    \addplot[myorange!80, line width=1.2pt] plot coordinates {
        (0,3132)
        (0.77,3050)
        }; 
        \addlegendentry{$\rho=5\%$}
        
       \nextgroupplot[ymin=0,ymax=700,
                       ytick={0,200,400,600},
                       axis x line*=bottom,
                       % xlabel=\footnotesize{Length of Sequence},
                       width=0.46\textwidth,
                       height=0.21\textwidth]
                       
    \addplot[myblue!80, line width=1.2pt] plot coordinates {
        (0.95,700)
        (1,691.984)
        (3,541) 
        (5,438)
        (7,415)
        (9,394)
        };    
    \addplot[myorange!80, line width=1.2pt] plot coordinates {
        (0.77,700)
        (1,610.2853)
        (3,426)
        (5,322) 
        (7,308)
        (9,293)
        };     
        \node[below,font=\scriptsize] at (axis cs:1,610.2853) {81\%};
        \node[above,font=\scriptsize] at (axis cs:3.1,545) {83\%};
        \node[above,font=\scriptsize] at (axis cs:5.055,439) {86\%};
        \node[below,font=\scriptsize] at (axis cs:3.055,426) {86\%};
        \node[above,font=\scriptsize] at (axis cs:7.055,415) {87\%};
        \node[above,font=\scriptsize] at (axis cs:9.1,394) {88\%};
        \node[below,font=\scriptsize] at (axis cs:5.03,322) {90\%};
        \node[below,font=\scriptsize] at (axis cs:7.045,308) {91\%};
        \node[below,font=\scriptsize] at (axis cs:9.1,303) {92\%};
        \end{groupplot}
        }
    \end{tikzpicture}
    \caption{Different length of sequence.}
   \label{fig:sequences lenth}
\end{figure}

\paragraph{Sequence Length}
To examine the influence of sequence length on attack effectiveness, we consider sequences of 1, 3, 5, 7, and 9 consecutive time steps. 
Our attack manifests in two variants for different sequence lengths: at length 1, it resembles a targeted method for single-time steps; at greater lengths, it serves as a perturbation method for the entire data. 
For stealth considerations, we avoid longer sequences, and unless otherwise stated, set the sequence length to 5 in all other experiments.
The results in Figure \ref{fig:sequences lenth} show that longer sequences lead to better attack outcomes.
This is attributed to longer sequences that can capture more complex decision patterns, which are likely to be more prominent and critical in the dataset.
In contrast, the decision patterns captured by shorter sequences become increasingly simple and transient. These brief sequences might only encompass a fraction of the agent's decision, resulting in a relatively minor impact on the agent's performance when attacked.

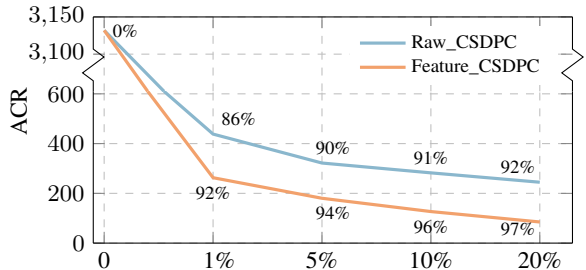
\begin{figure}[t!] % 使用figure环境来确保表格和图片都能被正确引用          
    \centering  
    \begin{tikzpicture}   
    \footnotesize{
        \begin{groupplot}[
        ymajorgrids,
        xmajorgrids,
        grid style=dashed,
        legend style={
        draw=none,
        line width=1pt,
        },
        legend style={at={(0.96,0.9)}, anchor=north east, legend cell align=left,
        nodes={scale=0.8, transform shape}},
        xmode=normal,
        group style={
                group name=my fancy plots,
                group size=1 by 2,
                xticklabels at=edge bottom,
                vertical sep=0pt
            },
        xmin=-0.01,xmax=0.43,
        xtick distance=1,
        xtick={0,0.1,0.2,0.3,0.4},
        xticklabels={0, \phantom{3}1\%, 5\%, 10\%, 20\%},
        xlabel style={yshift=0.0em},
        ]
        
        \nextgroupplot[ymin=3050,ymax=3150,
                       ytick={3100,3150},
                       axis x line*=top, 
                       ylabel style={at={(-0.12, -0.26)}}, 
                       ylabel=\footnotesize{ACR},
                       axis y discontinuity=crunch,
                       width=0.46\textwidth,
                       height=0.15\textwidth]
        \addplot[myblue!80, line width=1.2pt] plot coordinates {
             (0,3132) 
             (0.055,3050) 
            };
            \addlegendentry{Raw\_CSDPC}
        \addplot[myorange!80, line width=1.2pt] plot coordinates {
             (0,3132) 
             (0.04,3050) 
            };
            \node[right,font=\scriptsize] at (axis cs:0,3132) {0\%};
            \addlegendentry{Feature\_CSDPC}
        
       \nextgroupplot[ymin=0,ymax=610,
                       ytick={0,200,400,600},
                       axis x line*=bottom,
                       % xlabel=\footnotesize{Poisoning Rate},
                       width=0.46\textwidth,
                       height=0.21\textwidth]
        \addplot[myblue!80, line width=1.2pt] plot coordinates {
        (0.055,610)
        (0.1,438.2528) 
        (0.2,321.9049)
        (0.3,282.3755)
        (0.4,244.7133)
        }; 
        \addplot[myorange!80, line width=1.2pt] plot coordinates {
        (0.04,610)
        (0.1,262.8584) 
        (0.2,180.2488)
        (0.3,126.8107)
        (0.4,84.9307)
        }; 
        \node[right,font=\scriptsize] at (axis cs:0.1,500) {86\%};
        \node[above,font=\scriptsize] at (axis cs:0.21,322) {90\%};
        \node[above,font=\scriptsize] at (axis cs:0.3,282) {91\%};
        \node[above,font=\scriptsize] at (axis cs:0.38,245) {92\%};
        \node[below,font=\scriptsize] at (axis cs:0.1,260) {92\%};
        \node[below,font=\scriptsize] at (axis cs:0.21,180) {94\%};
        \node[below,font=\scriptsize] at (axis cs:0.3,127) {96\%};
        \node[below,font=\scriptsize] at (axis cs:0.38,115) {97\%};
        \end{groupplot}
        }
    \end{tikzpicture}
    \caption{Different poisoning rate.}
    \label{fig:poisoning_rate}
\end{figure}

\paragraph{Poisoning Rate}
To delve deeper into the impact of poisoning rates on attack effectiveness, we conduct the CSDPC attack with varying poisoning rates against the CQL agent in the Walker2D environment.
As shown in Figure \ref{fig:poisoning_rate}, the trend of average cumulative rewards varying with poisoning rates is depicted, $\eta =0.05$.
A clear trend emerges from the graph: as the poisoning rate $\rho$ increases, the average cumulative reward obtained by the agent under CSDPC attacks significantly decreases.
This result aligns with our expectation, as a higher poisoning ratio signifies the disappearance of more rare decision patterns in the dataset, and the data-policy coverage is significantly reduced.
Consequently, this leads to greater degradation in the agent's performance.

\begin{figure}[t]
\centering
\centering
\captionof{table}{Effect of cluster number $k$, $\rho=1\%$.}
\label{tab:cluster_k_ablation}
\renewcommand{\arraystretch}{1.2}
\footnotesize
\begin{tabular}{c|c|c|c|c}
\toprule
Environment & Algorithm & Clean & $k$ & ACR(AER) \\
\hline
\multirow{3}{*}{Walker2D} & \multirow{3}{*}{CQL} &\multirow{3}{*}{3132} & 6   & 689(78\%) \\
&&& 8   & 438(86\%) \\
&&& 10  & 626(80\%) \\
\bottomrule
\end{tabular}
\end{figure}

\paragraph{Effect of cluster number $k$}
\label{Effect of cluster number $k$}
To evaluate the impact of the number of clusters used for discretizing state-action pairs, we conducted an ablation study with raw data. 
We compare three different settings for the number of clusters: $k=6$, $k=8$ (selected using the elbow method), and $k=10$. 
The results in the Table \ref{tab:cluster_k_ablation} show that selecting $k$ via the elbow method leads to the most effective attack performance, likely due to a better trade-off between merging similar behaviors and preserving discriminative structure.

\section{Discussion}
In this paper, during the process of discretizing the state-action space, we prioritize computational efficiency for the attack, opting not to utilize a more refined method. 
This choice may have constrained the attack's effectiveness. 
In the future, we aim to develop more effective methods.

Our theoretical analysis based on the sequence-level concentrability coefficient assumes independence across steps, thus it does not explicitly account for correlated perturbations.
Such temporal correlations may amplify coverage deficiencies beyond the provided theoretical bounds.
Future work will explore modeling correlated biases across time steps to achieve tighter theoretical guarantees.

While this paper focuses on identifying and analyzing vulnerabilities, we acknowledge that robust defense mechanisms against sequence-level poisoning in offline RL remain largely unexplored. Current defense strategies in online RL rely on active environment interaction, which is infeasible in offline settings. Detection-based methods adapted from other domains also exhibit limited effectiveness in our evaluations. Motivated by these challenges, we propose several preliminary directions for defense:
\begin{itemize}
\item Coverage-Aware Data Augmentation: Leveraging generative models to synthesize trajectories in underrepresented regions can reduce distributional mismatch and mitigate poisoning effects.
\item Sequence Consistency Checking: Incorporating plausibility-based filtering during training can suppress anomalous transitions and reinforce learning from coherent behavioral patterns.
\item Robust Policy Optimization: Enhancing existing algorithms with sequence-level uncertainty regularization may reduce sensitivity to rare or corrupted trajectories.
\end{itemize}
We hope this work fosters research on secure and coverage-aware offline RL.

\section{Conclusion}

This paper provides an in-depth analysis and systematic empirical study of the robustness and sensitivity of offline RL when faced with data issues with insufficient sequence-level data-policy coverage. 
Theoretical analysis reveals that insufficient sequence-level coverage exponentially amplifies the upper bound of estimation errors, leading to significant performance degradation in policy learning.
We propose a collapsing sequence-level data-policy poisoning attack, thereby unveiling the potential impacts of distributional shift on offline RL performance.  
The goal is to inspire future researchers to collect high-quality datasets during the online phase and to further explore the design of robust and secure offline RL algorithms.

\section*{Acknowledgments}
We would like to thank anonymous reviewers for their constructive comments.
This work was supported by the National Natural Science Foundation of China (No.62406086, No.62272127), the Joint Funds of the National Natural Science Foundation of China (No.U22A2036, No.U21B2019), Basic Research Projects of the Central Universities and Colleges (3072025ZN0602), and Natural Science Foundation of Heilongjiang Province of China (No.TD2022F001).

\bibliography{example_paper}

%%%%%%%%%%%%%%%%%%%%%%%%%%%%%%%%%%%%%%%%%%%%%%%%%%%%%%%%%%%%
\newpage
\onecolumn

\title{Collapsing Sequence-Level Data-Policy Coverage \\
via Poisoning Attack in Offline Reinforcement Learning\\(Supplementary Material)}
\maketitle
\appendix

\section{CSDPC Algorithm}
\label{CSDPC Algorithm}
Algorithm \ref{CSDPC} describes the CSDPC attack process. The algorithm takes inputs that include the clean offline RL dataset $D$, the number of clusters $k$, the length $l$ of the sequence, the perturbation $\eta$, and the poisoning rate $\rho$. It produces a poisoned dataset $D'$. Output poisoned dataset $D'$.

\begin{algorithm}
\caption{CSDPC Attack Workflow}
\label{CSDPC}
\begin{algorithmic}
\STATE {\bfseries Input:} $(s,a)$: state-action pairs in clean offline RL dataset $D$, $k$: number of clusters, $l$: length of sequences, $\rho$: the poisoning rate
\STATE {\bfseries Output:} Poisoned dataset $D'$

\STATE Perform k-means clustering on all $(s,a)$ in dataset $D$ to partition the state-action space into $k$ clusters
\FOR{each $(s,a)$ in $D$}
    \STATE Assign a cluster label based on the k-means result
\ENDFOR
\STATE Initialize an empty dictionary for sequences cluster combinations and counts
\FOR{each sequence of length $l$ in $D$ starting at time step $t$}
    \STATE Record the sequence of cluster labels for the sequence $\tau_t = u_t, u_{t+1}, \ldots, u_{t+l}$
    \STATE Remove consecutive duplicate labels and obtain the decision pattern $p$
    \STATE Increment the count of the decision pattern in the dictionary
\ENDFOR

\STATE Identify cluster combinations with the lowest counts as critical sequences set $\mathcal{P}$ based on $\rho$
\FOR{each sequence in $\mathcal{P}$}
    \FOR{each $(s,a)$ in the sequence}
        \STATE Apply perturbations to $(s, a)$ to change the decision pattern to the most frequent decision pattern
    \ENDFOR
\ENDFOR
\STATE Combine all modified and unmodified sequences to form the poisoned dataset $D'$

\STATE \textbf{return }$D'$
\end{algorithmic}
\end{algorithm}

\begin{table*}[t!]
\caption{The hyperparameter settings for algorithms.}
\label{parameters}
\begin{center}
\footnotesize
\renewcommand{\arraystretch}{1.5} %调整行距
\begin{tabular}{ccccc}
\toprule
\multicolumn{1}{c}{Parameters} & \multicolumn{1}{c}{BCQ} & \multicolumn{1}{c}{BEAR} & \multicolumn{1}{c}{CQL} & \multicolumn{1}{c}{BC} \\ 
\hline
Optimizer&Adam&Adam&Adam&Adam\\ 
Critic Network Learning Rate&0.001&0.003&0.003&-\\ 
Actor Network Learning Rate&0.001&0.001&0.001&-\\ 
VAE Learning Rate&0.001&0.003&-&-\\ 
Batch Size&100&256&256&100\\ 
$\lambda$&0.75&-&-&-\\ 
Critic Network Hidden Units&[400, 300]&-&[256, 256, 256]&-\\ 
Actor Network Hidden Units&[400, 300]&-&[256, 256, 256]&-\\ 
VAE Encoder Hidden Units&[750, 750]&[750, 750]&-&-\\ 
VAE Decoder Hidden Units&[750, 750]&[750, 750]&-&-\\ 
$\gamma$&0.99&0.99&0.99&1\\ 
Activation Function&ReLU&ReLU&ReLU&ReLU\\ 
$\alpha$ Learning Rate&-&0.001&-&0.001\\ 
$\alpha$ Threshold&-&0.05&-&-\\ 
\bottomrule
\end{tabular}
\end{center}
\end{table*}

\section{Detailed Experimental Settings}
\label{Experimental Settings}
\subsection{Environment and Datasets}
\label{Environment and Datasets}

Our experiments were conducted on a computer with dual 12-core Intel(R) Xeon(R) CPUs (32GB RAM) and NVIDIA 3090. For a comprehensive evaluation of our attack framework, we selected four complex continuous tasks from the offline environment D4RL \citep{Fu2020D4RLDF}: Walker2D, Hopper, and Half-Cheetah in the MuJoCo robot simulator \citep{Todorov2012MuJoCoAP}, and Carla-Lane autonomous driving tasks in the Carla simulator \citep{Dosovitskiy2017CARLAAO}, which are closer to real-world scenarios.

The MuJoCo tasks require the agent to move forward rapidly in different forms. Each dataset is medium-level and comprises 1 million steps. The Carla-Lane task aims to drive a car smoothly and swiftly forward, with a dataset size of 100,000 pieces.

\subsection{Experimental Parameters}

In the Mujoco environment, the random seed is set to 0, in the Carla environment, it is set to 5.
The hyperparameter settings for the algorithms used in the experiments are presented in Table \ref{parameters}.

\subsection{Offline RL Algorithms}

To evaluate how offline RL algorithms perform under our poisoning framework, we employed three algorithms for training agent: Batch-Constrained Q-learning (BCQ) \citep{Fujimoto2018BCQ}, Batch-Ensemble Actor-Critic with Retrace (BEAR) \citep{Kumar2019StabilizingOQ}, Conservative Q-Learning (CQL) \citep{Kumar2020CQL} and Behavioural Cloning (BC) \citep{Sutton2018ReinforcementL}. We use the official open-source code of these algorithms and follow the settings by D4RL \citep{Fu2020D4RLDF}.

\begin{table*}[t!]
  \centering
  \caption{The values are the ACR obtained by the agent in various environments, the values in parentheses are the AER in poisoned agents, and $\eta = 0.05$, $\rho=5\%$.}
  \label{4environments_table_5}
  \renewcommand{\arraystretch}{1.3}
  \resizebox{\textwidth}{!}{
  \huge
  \begin{tabular}{c|rrr|rrr|rrr|rrr}
    \toprule
     \multirow{2}{*}{Algorithms} &\multicolumn{3}{c|}{Walker2D} &\multicolumn{3}{c|}{Hopper}&\multicolumn{3}{c|}{Half}&\multicolumn{3}{c}{Carla}\\ 
    \cline{2-13} 
    &\multicolumn{1}{c}{Clean}&\multicolumn{1}{c}{Raw}&\multicolumn{1}{c|}{Feature}&\multicolumn{1}{c}{Clean}&\multicolumn{1}{c}{Raw}&\multicolumn{1}{c|}{Feature}&\multicolumn{1}{c}{Clean}&\multicolumn{1}{c}{Raw}&\multicolumn{1}{c|}{Feature}&\multicolumn{1}{c}{Clean}&\multicolumn{1}{c}{Raw}&\multicolumn{1}{c}{Feature} \\ 
    \hline
    CQL & 3132 & 322(90\%) & 180(94\%) & 3158 & 180(94\%) &  70(98\%) & 4822 & 573(88\%) & 437(91\%) & 191 & 39(79\%)  & 24(87\%)\\
    BEAR& 2593 &  55(98\%) &  40(98\%) & 2119 & 111(95\%) &  66(97\%) & 4290 & 382(91\%) & 337(92\%) & 89  & 14(84\%)  &  8(91\%)\\
    BCQ & 2341 & 113(95\%) &  69(97\%) & 2823 & 238(92\%) & 113(96\%) & 4694 & 676(86\%) & 576(88\%) & 466 & 79(83\%)  & 53(89\%)\\ 
    BC  & 744  &  81(89\%) &  39(97\%) & 3450 & 229(93\%) &  99(97\%) & 4017 & 401(90\%) & 338(92\%) & 384 & 69(82\%)  & 56(85\%)\\
    \cline{1-13} 
    \multirow{2}{*}{Average} & 2203 & 143(93\%) & 82(96\%) & 2613 & 190(93\%) & 87(97\%) & 4456 & 508(89\%) & 421(91\%) & 283 & 50(82\%) & 35(88\%)\\
    \cline{2-13} 
     & \multicolumn{6}{c}{Raw Data AER \quad 89\%} & \multicolumn{6}{c}{Feature Data AER \quad 93\%}\\
    \bottomrule    
  \end{tabular}}
\end{table*}

\begin{table*}[t!]
\caption{AER of the poisoned agents under different attacks. The best results are indicated in bold.}
\label{compare table}
\begin{center}
\footnotesize
\begin{tabular}{cccccc}
\toprule
\multicolumn{1}{c}{Environments} & \multicolumn{1}{c}{Attack Methods} & \multicolumn{1}{c}{RL} & \multicolumn{1}{c}{Target Algorithms} & \multicolumn{1}{c}{$\eta$} & \multicolumn{1}{c}{AER} \\ 
\hline
\multirow{5}{*}{Hopper} & RS \citep{Zhang2020RobustDR}    &online& PPO & 0.75 &75\%\\
& SA-RL\citep{Zhang2020RobustDR} &online& PPO & 0.75 &80\%\\
& PA-AD \citep{SunZLH22}          &online& PPO & 0.75 &\textbf{95\%}\\
& BAFFLE \citep{gong2022mind}     &offline& CQL &  -   &48\%\\
& CSDPC (Ours)                     &offline& CQL & 0.05 &\textbf{97\%}\\ 
\hline
\multirow{5}{*}{Half-Cheetah} & RS \citep{Zhang2020RobustDR}    &online& PPO & 0.15 &93\%\\
& SA-RL \citep{Zhang2020RobustDR} &online& PPO & 0.15 &100\%+\\
& PA-AD \citep{SunZLH22}          &online& PPO & 0.15 &\textbf{100\%+}\\
& BAFFLE \citep{gong2022mind}     &offline& CQL &  -   &59\%\\
& CSDPC (Ours)                     &offline& CQL & 0.05 &91\%\\ 
\hline
\multirow{5}{*}{Walker2D} & RS \citep{Zhang2020RobustDR}    &online& PPO & 0.05 & 70\%\\
& SA-RL \citep{Zhang2020RobustDR} &online& PPO & 0.05 & 76\%\\
& PA-AD \citep{SunZLH22}          &online& PPO & 0.05 & 82\%\\
& BAFFLE \citep{gong2022mind}     &offline& CQL &  -   & 55\%\\
& CSDPC (Ours)                     &offline& CQL & 0.05 & \textbf{96}\%\\ 
\bottomrule
\end{tabular}
\end{center}
\end{table*}

\section{More Results}
\label{More Results}

\subsection{The Effectiveness of CSDPC Attack}
\label{The Effectiveness of CSDPC}

When the poisoning rate is 5\%, the effectiveness of CSDPC attack is as shown in Table \ref{4environments_table_5}.

\subsection{Additional Comparative Results}
\label{Additional comparative results}

Although the online attack method cannot be applied to offline RL due to the need to interact with the environment in real-time to obtain training parameters, however, in order to more fully evaluate the performance of the CSDPC method, we provide a succinct comparison between CSDPC attack and previously established online and offline RL attack methods. 
The comparative results are summarized in Table \ref{compare table}. 
The results demonstrate that the effectiveness of our attack method is optimized when the size of perturbations is identical or nearly so.  

In the Half-Cheetah environment, the magnitude of perturbations employed by other online RL attack methods is threefold that of our CSDPC attack. This substantial difference in perturbation magnitude could account for the capacity of these methods to not only markedly degrade the performance of the agents but also, induce penalties.

\end{document}